\documentclass{article}

     \usepackage[preprint]{nips_2018}

\usepackage[utf8]{inputenc} 
\usepackage[T1]{fontenc}    
\usepackage{hyperref}       
\usepackage{url}            
\usepackage{booktabs}       
\usepackage{amsfonts}       
\usepackage{nicefrac}       
\usepackage{microtype}      
\usepackage{amsfonts,amsmath,amssymb,array}
\usepackage{makeidx}         
\usepackage{graphicx}        
\usepackage{subcaption}
\usepackage{mwe}                             
\usepackage{multicol}        
\usepackage[bottom]{footmisc}

\usepackage{times}
\usepackage{parskip}
\usepackage{wrapfig}

\usepackage{amssymb}
\usepackage{epsfig}
\usepackage{amsfonts,amsmath,amssymb,array}
\usepackage{tabularx}   
\usepackage{colortbl}
\usepackage{longtable}
\usepackage{booktabs}
\usepackage{url}
\usepackage{icomma}
\usepackage{pstricks}
\usepackage{longtable}
\usepackage{multirow}
\usepackage{color}
\usepackage{algorithm2e}

\usepackage{subfiles}
\usepackage{float}
\usepackage{mathtools}

\title{Learning the Non-linearity in Convolutional Neural Networks}

\author{%
  Gavneet Singh Chadha\\
  Automation Technology Group\\
  South Westfalia University of Applied Science\\
  Soest, Germany \\
  \texttt{chadha.gavneetsingh@fh-swf.de} \\
 \And
  Andreas Schwung \\
	Automation Technology Group\\
  South Westfalia University of Applied Science\\
  Soest, Germany \\
  \texttt{schwung.andreas@fh-swf.de} \\
}

\begin{document}

\maketitle

\begin{abstract}
We propose the introduction of nonlinear operation into the feature generation process in convolutional neural networks. This nonlinearity can be implemented in various ways. First we discuss the use of nonlinearities in the process of data augmentation to increase the robustness of the neural networks recognition capacity. To this end, we randomly disturb the input data set by applying exponents within a certain numerical range to individual data points of the input space. Second we propose nonlinear convolutional neural networks where we apply the exponential operation to each element of the receptive field. To this end, we define an additional weight matrix of the same dimension as the standard kernel weight matrix. The weights of this matrix then constitute the exponents of the corresponding components of the receptive field. In the basic setting, we keep the weight parameters fixed during training by defining suitable parameters. Alternatively, we make the exponential weight parameters end-to-end trainable using a suitable parameterization. The network architecture is applied to time series analysis data set showing a considerable increase in the classification performance compared to baseline networks.   
\end{abstract}

\section{Introduction}

Convolutional Neural Networks~\citep{Fuku1982,LeCun1989} are known to provide superior performance not only in image recognition tasks but also in speech recognition, natural language processing and time series analysis. The operation of one layer consist of a certain number of channels where a corresponding filter with trainable weights is convolved with the input. The convolution itself is basically a linear operation while the weights to be learned can take arbitrary weight. The nonlinearity of the network is introduced by different nonlinear activation function like ReLu or sigmoids in each layer channel. 

In this work we propose extensions to the classical CNN architecture with a focus on applying nonlinear operations to the input data as well as to the individual layer inputs in different ways. The motivation behind this lies in the fact that the underlying process in many applications in machine learning appears to be nonlinear to a certain degree. In fact, in time series analysis, a lot of technical processes have a strongly nonlinear behavior. In such cases, the representational power of CNNs, especially when using the popular ReLu-functions, is somewhat limited. Hence, introducing nonlinearity in the form of exponents can increase the capability and includes an additional form of inductive bias to the CNN. Note that exponentials are a natural component of series expansions which can represent general nonlinear functions with arbitrary precision. In addition, introducing nonlinearities in the network can also help to flatten the learning manifolds for the weights in such a way, that it is less prune to getting stuck in local minima.We show an example later in the text.

As a first approach, we experiment with nonlinear data augmentation techniques. Data augmentation is well known for providing additional robustness to the training performance. Examples includes translation, scaling, cropping or mirroring~\citep{Huang2017} in image recognition and window slicing and warping~\citep{Guennec2016} in time series analysis. We follow a similar approach by disturbing data points of the input using exponents, i.e. we randomly assign an exponent to the data point. We choose the range of the exponents between -2 and 4 with uniform sampling. 

Although the nonlinear data augmentation already improves the performance of the CNN as will be shown later, the approach possess some inherent weakness. As we do the augmentation in the input space, the exponent of each data point is fixed a priori using random sampling. Hence, suboptimally sampled exponents can even degrade the performance of the network considerably. To generalize this operation we employ the nonlinear operation in the convolution operation itself leading to nonlinear convolutions. More specific, we define an additional weight matrix, named exponent weight matrix (EWM), with dimension equal to the dimension of the receptive field. Then we use the EWM to element-wise assign an exponent to the component of the receptive field. As with the standard weights in CNN, the EWM is shared for each neuron. With this architecture, we can either fix the EWM a priori by assigning meaningful weight patterns for the receptive field. Alternatively, we can make the EWM end-to-end trainable.

We experiment on a variety of different machine learning task including image classification and time series analysis and show the results.

\section{Nonlinear Convolutional Neural Networks}

We propose different ways to incorporate nonlinear data operations into the CNN framework. On the hand we show that nonlinear data augmentation of the input space can lead to improved results in time series analysis. On the other hand, we apply nonlinear operations to the convolutional layers of the CNN.

\subsection{Nonlinear Data Augmentation for Time Series Analysis}

\begin{figure}
        \centering
        \begin{subfigure}[b]{0.475\textwidth}
            \centering
            \includegraphics[width=\textwidth]{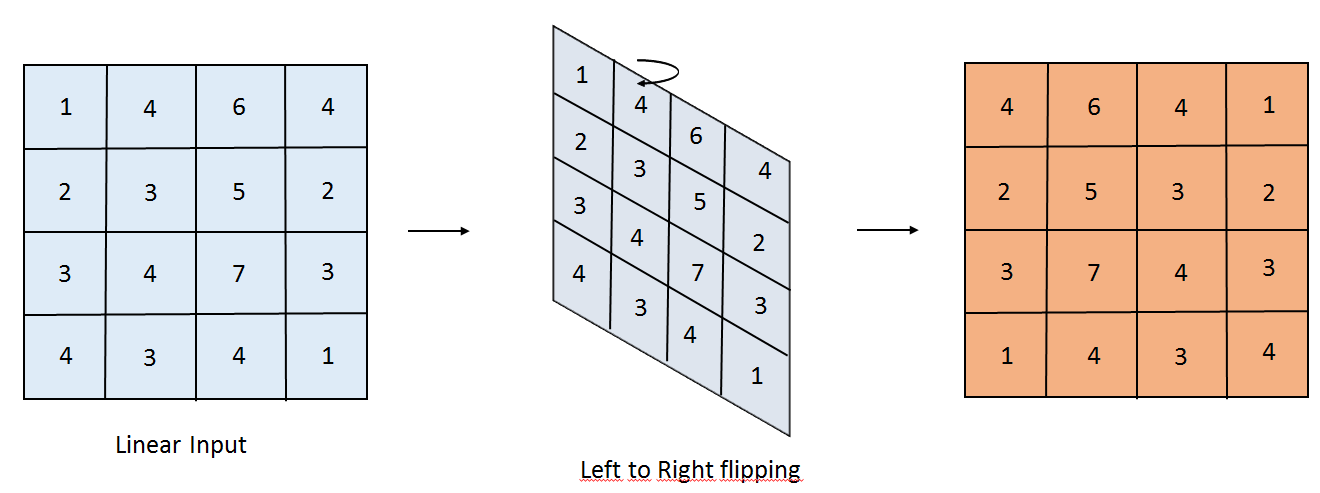}
            \caption[Network2]%
            {{\small Left to Right Flipping.}}    
            \label{fig:l2r}
        \end{subfigure}
        \hfill
        \begin{subfigure}[b]{0.475\textwidth}  
            \centering 
            \includegraphics[width=\textwidth]{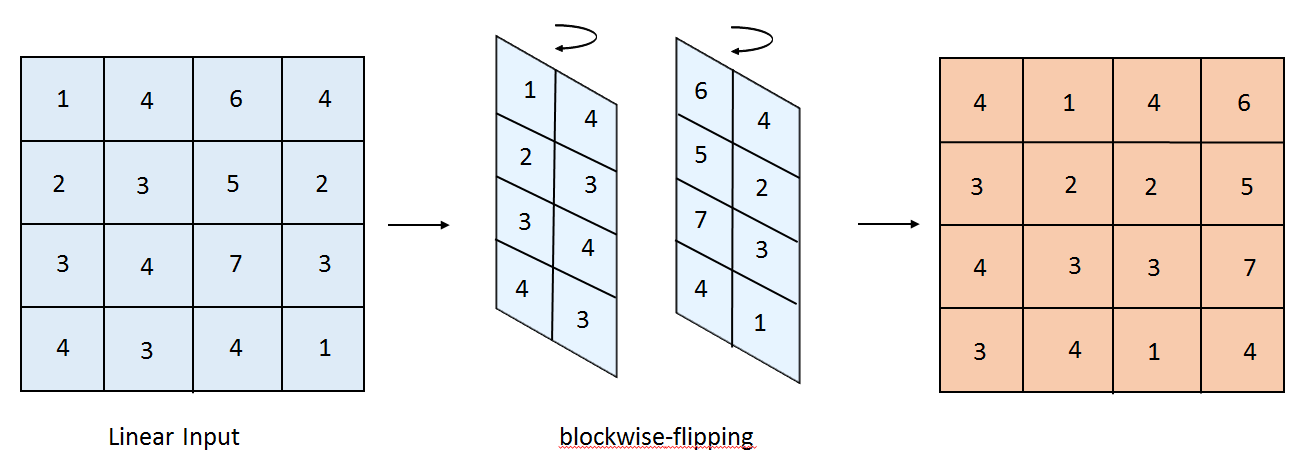}
            \caption[]%
            {{\small Blockwise Flipping.}}    
            \label{fig:bw}
        \end{subfigure}
        \vskip\baselineskip
        \begin{subfigure}[b]{0.475\textwidth}   
            \centering 
            \includegraphics[width=\textwidth]{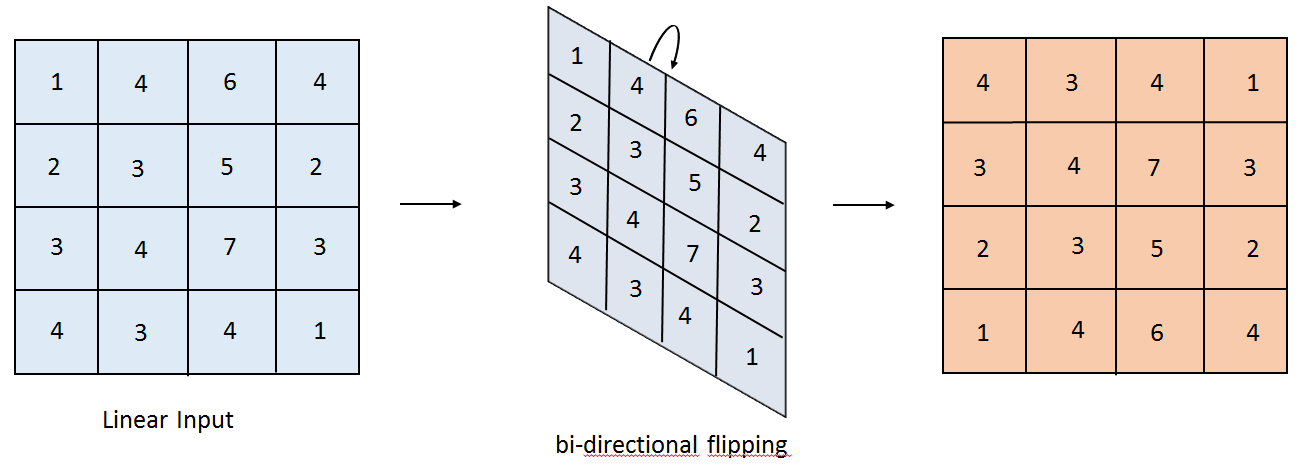}
            \caption[]%
            {{\small Bi-Directional Flipping}}    
            \label{fig:bidi}
        \end{subfigure}
        \quad
        \begin{subfigure}[b]{0.475\textwidth}   
            \centering 
            \includegraphics[width=\textwidth]{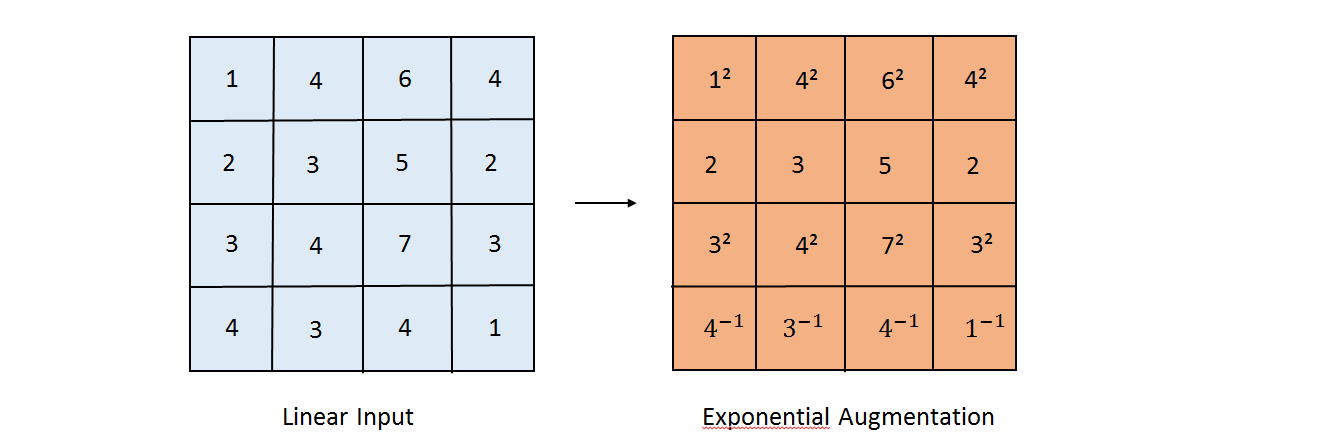}
            \caption[]%
            {{\small Exponential Augmentation}}    
            \label{fig:nl}
        \end{subfigure}
        \caption[ Proposed data perturbation methods for time series analysis tasks. ]
        {\small Proposed data perturbation methods for time series analysis tasks.} 
        \label{fig:dataaugmethods}
\end{figure}

A simple, yet effective way to improve the robustness and generalisation of time series analysis are data augmentation techniques. Such techniques are frequently applied in the image recognition context, where random noise as well as flipping on the pixel level is added to the input image. We propose a similar approach for time series analysis with different forms of perturbations and analyze the impact on the performance of training. Particularly, we use various perturbations like Left to Right flipping, Blockwise flipping, Bi-directional flipping and exponential assignments to input rows as illustrated in Fig.~\ref{fig:dataaugmethods}. 
The results are illustrated in Sec.~\ref{sec:exp}.

\subsection{Nonlinear Convolutional Neural Networks}

We start with the standard convolution operation. We consider an input patch, also called receptive field, $X \in \mathbb{R}^{k_h\times k_w}$ with $n = k_h\cdot k_w$ elements (normally $k_h=k_w$), which can be reshaped as a vector $x\in \mathbb{R}^{n}$. Also consider a weight matrix $W_1 \in \mathbb{R}^{k_h\times k_w}$ which can similarly be reshaped as a vector $w_1\in \mathbb{R}^{n}$. Then, the convolved output of this linear filter yields
\begin{align}\label{eq:conv}
	y(x) = x \ast w_1 = \sum^n_{i=1} w_{1,i} \cdot x_i + b
\end{align}
where $w_{1,i}$ denotes the $i$th element of $w_1$ and $b$ is a bias term.

As previously discussed we introduce additionally weight matrices to incorporate nonlinearity into the convolution operation. Particularly, we concentrate on providing the elements of the receptive field $I$ with exponents. 

\subsubsection{Exponential weight matrix}

We introduce the EWM $W_2 \in \mathbb{R}^{k_h\times k_h}$ which can similarly be reshaped as a vector $w_2\in \mathbb{R}^{n}$ and define the nonlinear convolution operation as
\begin{align}\label{eq:nlconv}
	y_{nl}(x) = x^{w_2} \ast w_1 = \sum^n_{i=1} w_{1,i} \cdot x^{w_{2,i}}_i + b.
\end{align}
where the exponential operation is considered element-wise. Eq.~\eqref{eq:nlconv} can be rewritten as~\citep{Trasky2018}
\begin{align}\label{eq:nlconvalternative}
	y_{nl}(x) = \exp^{\text{diag}(w_2) \cdot \log(x)} \ast w_1.
\end{align}
where the exponential and logarithmic operation are applied element-wise and $\text{diag}$ .

\textbf{Weight sharing:} 
The above EWM introduces additional parameters to the CNN which might increase the risk of overfitting if not appropriately defined. However, for different applications, a reduction of the number of weights can be achieved by suitable weight sharing. For instance, it might be beneficial to share the exponent in each row of the receptive field. In time series analysis this translate to an assigment of identical exponents to time steps. This can be achieved by defining
\begin{align}\label{eq:weightshare1}
	y_{nl}(x) = \exp^{\text{diag}(w^*_2 \otimes \mathbf{1}^{1\times k_h}) \cdot \log(x)} \ast w_1.
\end{align}
where the exponential and logarithmic operation are applied element-wise, $w^*_2\in \mathbb{R}^{k_h}$, $\mathbf{1}^{1\times k_h}$ denotes an all one vector of size $k_h$ and $\otimes$ denotes the Kronecker product.
Similarly, it might be beneficial to apply the same exponent to the elements of a sensor channel. In this case we obtain
\begin{align}\label{eq:weightshare2}
	y_{nl}(x) = \exp^{\text{diag}(\mathbf{1}^{1\times k_h} \otimes w_2) \cdot \log(x)} \ast w_1.
\end{align}
where the exponential and logarithmic operation are applied element-wise. Other patterns can be defined similarly.

\subsubsection{Generalized exponential weight matrix}

The above formulation is rather limited in the sense that just an exponent is applied to the element $x_i$ of the receptive field. Moreover, no relation between elements of the receptive field are explored. Hence, we extend the approach and apply more general interactions between the elements of the receptive field further increasing the capacity of the convolution operation. To this end we propose the following generalization to~\eqref{eq:nlconvalternative}
\begin{align}\label{eq:nlconv2}
	y_{nl}(x) = \exp^{W_3\cdot \log(X) \cdot W_4} \ast W_1.
\end{align}
where the exponential and logarithmic operation are applied element-wise and $W_3, W_4 \in \mathbb{R}^{k_h \times k_h}$. Alternative, we can set
\begin{align}\label{eq:nlconv3}
	y_{nl}(x) = \exp^{W_5 \cdot \log(x)} \ast w_1.
\end{align}
with $W_5 \in \mathbb{R}^{n\times n}$ and again element-wise exponential and logarithmic operation. The formulation in~\eqref{eq:nlconv3} provides the most general exponential relations at the cost of $n^2$ additional parameters. The formulation in~\eqref{eq:nlconv3} requires $2n$ and hence less parameters. However, the exponents to the elements of the receptive field are coupled. We compare the three approaches in the experiment section. Note that the above configuration of the nonlinear convolutions can be either fixed a priori or made end-to-end trainable which will be described next.

\subsection{End-to-end Training of Nonlinear Convolutional Neural Networks}\label{sec:e2e}

\begin{figure}[h]
 \centering
 \includegraphics[width=\columnwidth,keepaspectratio]{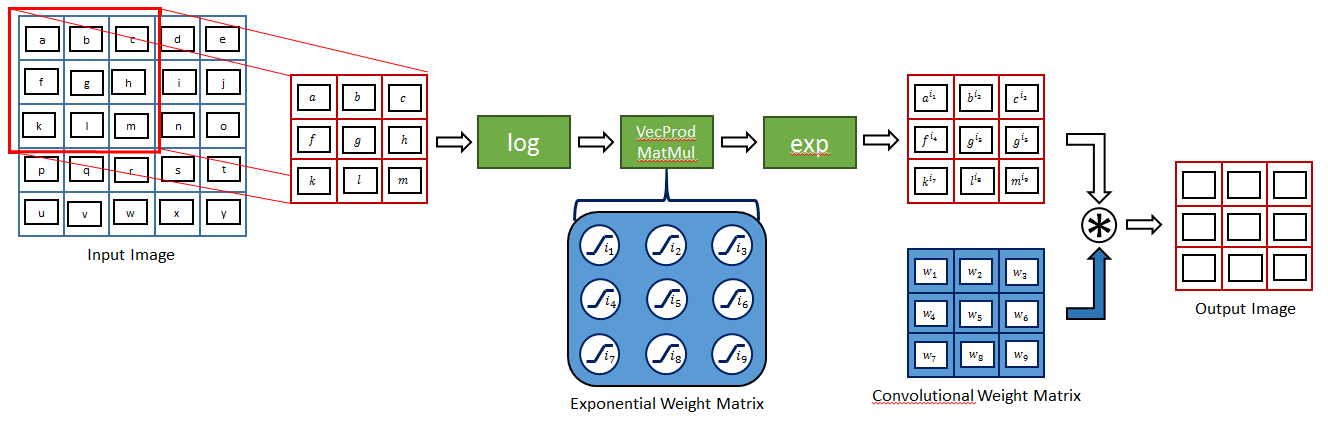}
\caption{Architecture of one unit of NLCNN for end-to-end training.}
\label{fig:architecture}
\end{figure}

As the proposed network architecture is fully differentiable with respect to the weight parameters, an end-to-end training procedure can be derived. Based on the previous section, the following feedforward path equations through the CNN layer $l$ can be derived:
\begin{align}
	\text{Nonlinear filter equation:}\quad & y^l_{nl}(x_l) = \exp^{W^{l,m} \cdot \log(x_l)} \ast w^{l,m}_1,\\
	\text{Output nonlinearity:}\quad & x_{l+1}=\sigma^l(y^l_{nl}(x_l)),\\
	\text{Weight parameter constraints:}\quad & v^l_{\min} < W^{l,m} < v^l_{\max},
\end{align}
where $\sigma^{l,m}$ is the activation function and $W^{l,m}$ is the exponential weight matrix of the $m$ neuron in the $l$th layer defined as in~\eqref{eq:nlconvalternative}-\eqref{eq:nlconv3}, respectively. The architecture of the NLCNN layer is illustrated in Fig.~\ref{fig:architecture}.
The weight parameters are constraint to avoid unreasonable low and high exponents. This constraints can be implemented either by parameter or gradient clipping such that the weight will not cross the limits. Alternatively, we can define unconstrained weights $\hat{W}^{l,m} \in \mathbb{R}$ which are subsequently send through an activation function $\sigma$
\begin{align}\label{eq:nlconv3}
	W^{l,m} = \sigma(\hat{W}^{l,m}).
\end{align}
However, the design of the activation functions deserves some deeper considerations. The main purpose of the activation function is to constrain the exponent weights to a predefined interval $[v_{\min}\ v_{\max}]$. Beside, we do not intend to overemphasize certain regions within this interval due to an unbalanced gradient of the activation function which is the case for sigmoidal activation functions which tend to push the parameter to the extremes. On the other hand, the activation function has to assure that the parameter can potentially recover from its maximal or minimal value. This is not assured by e.g. rectified linear units~\citet{Hahn2000} which cannot recover from values smaller than zero due to the zeroed gradient. In the experiments, we compare various types of activation functions for their suitability.


The standard weight matrix $W_1$ of CNNs are typically randomly initialized. However, such a random initialization within $[v_{\min}\ v_{\max}]$ appears to be unsuitable for the EWM $W_2$, $W_3$, $W_4$, $W_5$ due to the potentially high impact of the parameter. I.e. a random initialization near the maximal and minimal values has a considerably higher impact on the output results than the standard weight initialization. Hence, we propose to initialize the EWM $W_2$ as as an all one matrix while $W_3$, $W_4$ and $W_5$ are initialized as identity matrices. With this choice the NLCNN initializes as a standard CNN.   

\section{Related Work}

Recently, various forms of improvements to the conventional CNN has been made which considerably improve the performance of CNNs. Residual Networks (ResNets) have been introduced by~\citet{He2016}, incorporating shortcut connections in parallel to convolutional layers which reduces the vanishing/exploding gradient problem and allows for increasing the depth of the networks considerably. In~\citet{Huang2017} DenseNets are presented connecting each layer to every other depper layer,
which further mitigates the vanishing-gradient problem. \citet{Szegedy2015} introduced “Inception modules”, which uses filters of variable sizes in a parallel manner to capture different visual patterns of different sizes with further improvements presented in~\citet{Szegedy2017}. Similarly, ResNeXt~\citet{Xie2017} use repeating building blocks aggregating a set of transformations with the same topology. Spatial transformer networks~\citep{Jaderberg2015} inserts learnable modules to CNN for manipulating transformed data to assure invariance to spatial transformation. Dilation operation is proposed in~\citet{Yu2016} to cover broader spatial structures by blowing up the receptive field while keeping the weight matrix dimension fixed. A more general, but similar approach are deformable neural networks~\citep{Dai2017}. Recurrent neural filters containing a recurrent connection in the filter are proposed in~\citet{Yang2018}. However, all the above improvements are based on standard CNN operations and do not consider nonlinear operations in the receptive field.

Nonlinear receptive fields in form of quadratic forms are first analyzed and interpreted by~\citet{Berkes2006} as they follow some of the properties of complex cells in the primary visual cortex. Nonlinear convolutions have been introduced in the “Network in Network (NIN)” in~\citet{Lin2014}. In NIN, micro neural networks with more complex structures are used to abstract the data within the receptive field. For the input-output mapping, they use MLPs as a non-linear function approximator. The output feature maps are obtained by sliding the micro networks over the input in a similar manner as CNN. A special form of nonlinear convolution operations are employed in~\citet{Zoum2017}using a Volterra series model with second-order kernel representing the coefficients of quadratic interactions between two input elements. However, this introduces a large number additional parameters and increases the training complexity exponentially. Spline-CNN~\citet{Fey2018} are introduced with special emphasis on processing graph inputs by extending convolution operation by means of continuous B-spline basis functions parametrized by a constant number of trainable control values. Recently, kervolutional neural networks~\citet{Chen2019} are presented where the convolution is replaced by a nonlinear kernels on receptive field and weights with fixed structure. Parameters of the kernels are end-to-end trainable. However, none of these approaches also train the degree of nonlinearity as we do with training the EWM. Rather, the nonlinear operation on the input space has to be fixed a priori requiring for deeper knowledge on the problem while only the corresponding weights are trained. Furthermore, we keep the number of additional parameters introduced low in contrast to other approaches which allows for faster and more efficient training as well as better generalization. Furthermore, all the previous works focus on image recognition tasks while our focus is more on time series analysis and their specific challenges.

\section{Experiments}\label{sec:exp}

In this section, we present first results using the various proposed nonlinear convolution operation. We do the comparison on a benchmark time-series analysis data set collected from the Tennessee Eastman Process~\citep{Downs1993} employed for testing data-based fault diagnosis approaches. 

\subsection{Tennessee Eastman Process}\label{sec:TEP}

\begin{figure}[h]
 \centering
 \includegraphics[width=\columnwidth,keepaspectratio]{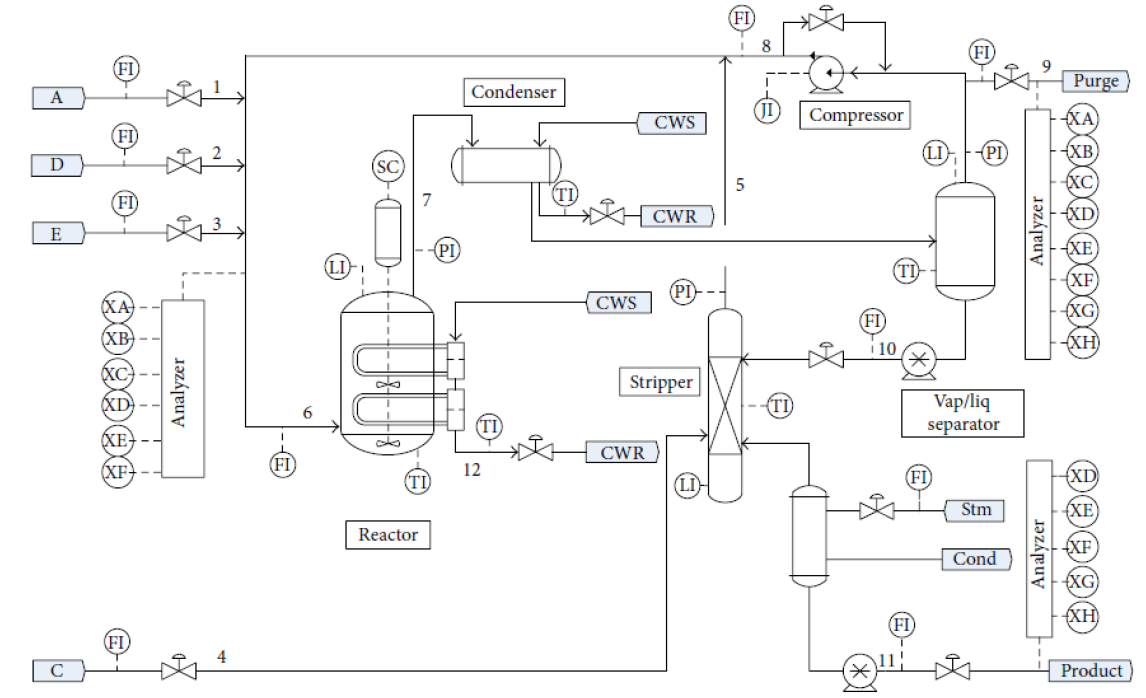}
\caption{Schematic of the Tennessee Eastman Process~\citep{Yin2012}.}
\label{fig:TE}
\end{figure}

The process simulation consists of five major units: a reactor, condenser, compressor, separator, and stripper. The process produces two products (G, H) from four reactants (A, C, D, E) with one byproduct (F) and one inert element (B). The process schematic is shown in Fig.~\ref{fig:TE}~\citep{Yin2012}.

Since the mathematical equations of the process are hardly to derive, TE process is an ideal system for evaluating data-driven techniques for the purpose of fault diagnosis. The TE process simulator has been widely used by researchers as a source of data for comparing various data based fault diagnosis methods~\citep{Yin2014,Kulkarni2005,Kano2002}. The datasets can be downloaded from http://web.mit.edu/braatzgroup/links.html. 

\begin {table}
\caption {Description of Process Faults in TE} \label{tab:title} 

\begin{tabular}{|l|c|r|}
	\hline
	\textbf{Faults}   &  \textbf{Description}   &   \textbf{Type} \\
	\hline
	1  &  A / C feed ratio (B composition constant)   &  Step Change   \\
	
	2  &  B composition (A / C feed ratio constant)  &  Step  Change   \\
	
	3  &  D Feed temperature  &  Step  Change  \\
	
	4  &  Reactor cooling water inlet temperature  &  Step Change    \\
	
	5  &  Condenser cooling water inlet temperature   &  Step  Change \\
	
	6  &  A Feed loss   &  Step Change   \\
	
	7  &  C header pressure loss    &  Step Change   \\
	
	8  &  A, B, C feed composition   &  Random  Variation  \\
	
	9  &  D feed temperature   & Random Variation  \\
	
	10  & C feed temperature   & Random Variation   \\
	
	11  & Reactor cooling water inlet temperature  & Random Variation   \\
	
	12  & Condenser cooling water inlet temperature  & Random Variation  \\
	
	13  & Reaction kinetics &  Slow drift   \\
	
	14  &  Reactor cooling water valve  &  Sticking    \\
	
	15  & Condenser cooling water valve  &  Sticking  \\
	
	16 - 20 &  Unknown  &  Unknown   \\ 
	
	21 &  The valve fixed at steady state position   & Constant position  \\
	\hline
	
\end{tabular}
\end {table}
The process provides the capability to measure 52 variables out of which 41 are process variables and the other 11 are manipulated variables. The dataset consists 22 training and 22 testing sets corresponding to each of the 21 process faults defined in~\citet{Chiang2001} and one normal operating condition. The process faults are described in Table~\ref{tab:title}.
Each faulty training set consists of 480 samples, all of which are faulty data samples. Meanwhile, each test set consists of 960 samples corresponding to 48 hours plant operation time, with the fault being introduced after the simulation time of 8 hours which corresponds to 160 samples. In a proprecessing step, the data is normalized to zero mean and unit covariance for each measured variable.

\subsection{Results}\label{sec:results}

Work on experimental results is ongoing and will be added when available.

\section{Conclusion}

We presented nonlinear convolutional neural networks, a novel CNN architecture which uses nonlinear operations on the input space of the CNN layer. We propose and compare three different settings for nonlinear operation in CNN. First we propose nonlinear data augmentation of the input space which can be seen as a preprocessing stage where we randomly assign exponent to the data points. Second we propose nonlinear convolutional neural networks where we define two weight matrices for each kernel operation, with one being the standard weight matrix while the other matrix defines the exponents applied to the receptive field. This exponents are fixed a priori. Finally, we propose an end-to-end training procedure where beside the weight matrix also the exponent weight matrix is trained. 


In future research we will apply and test NLCNN on various data sets, including image classification.

\end{document}